\documentclass{ifacconf}

\usepackage{graphicx}      
\usepackage{natbib}        
\usepackage{amsmath}
\usepackage{color}

\begin{document}
\begin{frontmatter}


\title{Learning Physics-Informed Surrogate Model of Linear Elastic Displacement Fields from Geometry}

\thanks[footnoteinfo]{This work benefited from French state aid managed by the National Research Agency under France 2030 with the reference ``ANR-22-PESP-0005''.}

\author[First]{Rodolphe Barlogis} 
\author[Second]{Ferhat Tamssaouet} 
\author[First]{Quentin Falcoz}
\author[First]{Stéphane Grieu}


\address[First]{PROMES-CNRS, Université Perpignan Via Domitia (UPVD), Perpignan, France}
\address[Second]{LAAS-CNRS, Université de Toulouse (UT), Toulouse, France}

\begin{abstract}                


This work aims to develop a fast and physically consistent surrogate model for real-time structural health monitoring of fractured elastic domains. We propose a physics-informed DeepONet framework that predicts displacement fields from both boundary conditions and fracture geometry, using a dedicated encoding strategy for the latter and without relying on finite-element-generated training data. The traction-free condition on the fracture boundary is imposed weakly through a localized penalty term. The presented numerical example focuses on one representative fracture geometry, demonstrating the feasibility of the formulation and laying the groundwork for extensions to surrogate modeling across diverse fracture geometries.
\end{abstract}


\end{frontmatter}

\section{Introduction}


Structural health monitoring and management is required in numerous industrial applications. Therefore, accurately and in real time predicting the mechanical response of structural components is essential for the next generation of industrial digital twins. The latter can be used to generate data to train models for design or monitoring purposes (e.g. fault detection and diagnosis). However, structural parts often have geometric discontinuities, such as holes, notches and cracks, which create strong spatial gradients and usually require high-fidelity finite element method (FEM) simulations. Although the FEM provides accurate solutions, the computational cost can be prohibitive in settings involving repeated evaluations across different geometries, loads or boundary conditions. This is typically the case in monitoring, control or design loops for smart industrial assets \citep{ciklamini2025enhancing}.

As a consequence, classical FEM solvers become a computational bottleneck in applications requiring repeated evaluations, motivating the development of fast surrogate models. This is where operator learning comes into play. Combined with automatic differentiation, operator learning appears suited for solving problems described by partial differential equations (PDEs). Rather than approximating pointwise input-output relationships, neural operators aim to learn the solution operator that maps functional inputs, such as distributed loads, boundary conditions, or geometric descriptors, to entire fields. For example, this could be displacement or strain fields over a domain. 


The present study proposes a physics-informed operator-learning methodology for predicting displacement fields in elastic solids subjected to mechanical loading and containing geometric gaps or localized internal features. More specifically, the objective is to investigate the ability of a DeepONet-based framework \citep{luDeepONetLearningNonlinear2021} to learn the mapping from boundary conditions and a geometry description to the full displacement field, without relying on FEM-generated training data. The main originality of the work lies in the proposed geometry-encoding strategy, which allows the fracture or internal void geometry to be provided directly as an input to the operator. This formulation makes the approach compatible with non-parametric geometry descriptions and enables geometry-aware learning of the mechanical response. While the numerical results presented in this paper focus on one representative fracture geometry, they serve as a proof of concept for the proposed formulation and illustrate its potential for extension to more diverse fracture shapes and configurations.

\begin{figure*}[!h]
    \centering
    \includegraphics[width=0.55\textwidth]{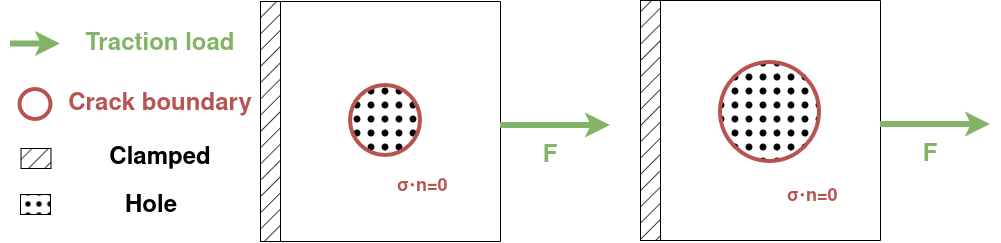}
    \caption{2D domains with central hole and the corresponding applied boundary conditions. }
    \label{fig:domain}
\end{figure*}


The paper is organized as follows: the first section provides a review of the existing literature related to the topic under consideration, highlighting recent contributions from the field of neural operators. Next, the proposed methodology is presented, distinguishing two possible options for solving a simple linear elastic problem. This is followed by an evaluation of the two options based on preliminary results. Finally, a section is provided for discussion and future perspectives.



\section{Related Work}








The numerical computation of displacement fields in linear elasticity is traditionally addressed through model-based approaches, where the governing equilibrium equations, constitutive laws, and boundary conditions are discretized and solved numerically. The finite element method (FEM) is the standard framework for this purpose, with well-established mathematical and algorithmic foundations for small-strain elasticity \citep{brennerMathematicalTheoryFinite2008}. These physics-based methods provide reliable solutions, but they require solving a discretized mechanical problem for each new loading case or geometry, which can become costly in many-query or real-time monitoring contexts.


The work of \cite{kiyaniPredictingCrackNucleation2025} is the closest to the approach proposed here to train a surrogate model. The authors also employ the DeepONet framework \citep{luDeepONetLearningNonlinear2021} to predict crack nucleation and propagation by using the phase-field method. This method treats fracture as an energy minimization problem, with the corresponding energy functional serving as the loss function that guides the training process.

In their formulation, the standard multi-layer perceptron (MLP)-based trunk network is replaced by a Kolmogorov–Arnold network (KAN), which enables a richer representation of spatial dependence. The resulting models are evaluated using several fracture scenarios, including nucleation, propagation and kinking.

In \cite{drosopoulosDeepONetPredictionFailure2025}, the authors propose a methodology based on DeepONets to predict the failure response of a 2D fiber-reinforced composite plate under various loading conditions. They constructed a dataset of nonlinear finite-element simulations over different domain configurations in order to train the neural operator, DeepONet. The loading conditions and fiber-reinforcement characteristics are both encoded as integers and supplied to the branch network, while the trunk network receives the spatial coordinates of the evaluation points. Unlike in \cite{kiyaniPredictingCrackNucleation2025}, both the branch and the trunk networks are implemented as fully connected architectures. The authors compare the predictive accuracy of the DeepONet model with that of a conventional feed-forward neural network, using the same input and output structures, but without the operator-learning decomposition. They also assess generalization on unseen combinations of branch and trunk inputs, demonstrating the improved predictive capability of the DeepONet framework.




Although these approaches demonstrate the potential of DeepONets for fracture mechanics, they still rely on datasets generated from finite-element simulations. In contrast, the present work investigates a fully physics-informed formulation trained without FEM-generated data, as detailed in the following section.

\section{Methodology}


This paper aims to predict the displacement field over a given domain. To this end, we provide the domain's geometry, physical properties, and boundary conditions as inputs to a model that learns to reconstruct the corresponding physical solution. To achieve this, the learning process of neural networks is guided by enforcing the governing partial differential equations of the problem, so that the predictions remain consistent with the underlying physics. The objective is to obtain a lightweight surrogate model with fast inference whose predictions depend explicitly on the domain geometry. This model should outperform standard FEM-based methods in terms of computational cost, making it suitable for applications requiring repeated solution evaluations.


The proposed method is illustrated in this paper through a simple example. Consider a 2D domain that is described by Fig. \ref{fig:domain}. The domain is fixed on the left, free on the top and bottom, and a force is applied to the right. Additionally, there is a hole located at the center that is enforced with a traction-free boundary condition. For simplicity, the fracture is modeled as a circular hole but with varying radii.


The mapping from the fracture geometry (in this case, the radius of a circle) of the 2D tensile-loaded domain to the corresponding physical solution is learned using a neural-operator approach. The stress field can subsequently be obtained from this solution.

\subsection{The DeepONet framework for a linear elastic problem.}

\begin{figure*}[ht]
    \centering
    \includegraphics[width=0.73\textwidth]{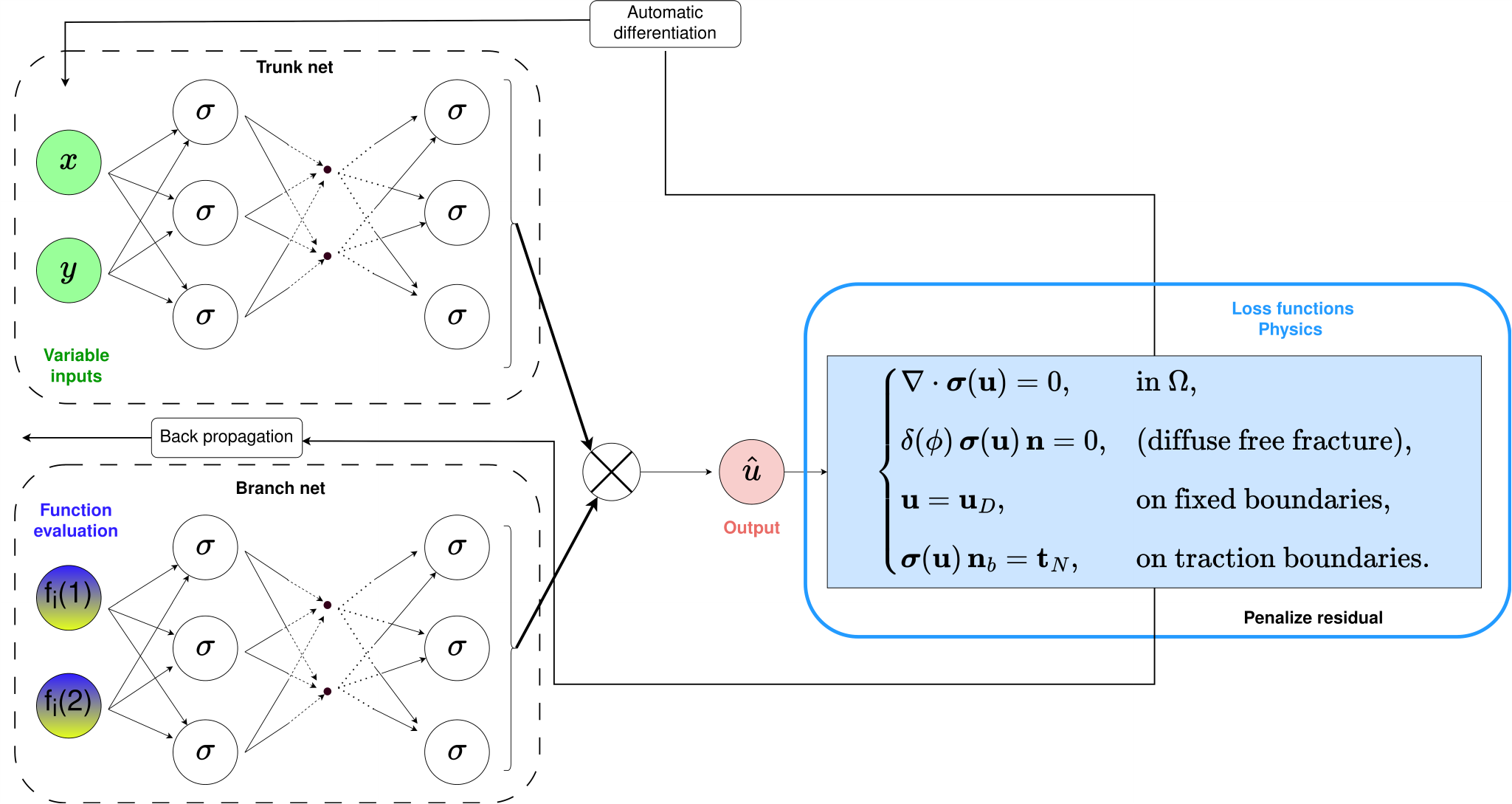}
    \caption{Schematic representation of the physics-informed DeepONet architecture used in this work.}
    \label{fig:deeponet}
\end{figure*}

Deep Operator Networks (DeepONets) \citep{luDeepONetLearningNonlinear2021} are neural architectures specifically designed to learn linear and nonlinear operators. Whereas classical neural networks approximate mappings between finite-dimensional vectors, DeepONets aim to approximate mappings from input functions to output functions—i.e., solution operators arising from parametrized differential equations or other functional input–output relationships.

A DeepONet consists of two subnetworks with distinct roles. The branch network processes a discrete representation of the input function, which is usually obtained by sampling it at a set of predetermined sensor points. The trunk network encodes the coordinates (or more generally, the query locations) at which the output function is evaluated. These two representations are then combined, most commonly via an inner product between the latent outputs of the branch and trunk networks, to produce the operator’s evaluation at the queried points.

This architectural decomposition has been shown to give DeepONets universal approximation capabilities for a broad class of operators, including those governing solutions of parametrized partial differential equations (PDEs) \citep{luDeepONetLearningNonlinear2021}. Consequently, DeepONets offer a systematic approach to learning high-dimensional solution maps dependent on boundary conditions, forcing terms, material distributions, and geometric parameters.

In the setting considered here, the operator is defined by the equations of linear elasticity for an isotropic solid, together with the geometric features that characterize the domain and the boundary conditions. These inputs fully determine the solution operator that maps boundary conditions and geometry to the displacement field. A schematic representation of this configuration, and of the manner in which it conditions the operator, is provided in Fig.~\ref{fig:deeponet}.


The DeepONet output is the displacement field $u$, consisting of its two components ($u_x$, $u_y$). Through automatic differentiation \citep{baydinAutomaticDifferentiationMachine2018a}, which enables efficient computation of spatial derivatives of the network output with respect to its inputs, the model can be constrained to satisfy the governing equations of linear elasticity at arbitrary evaluation points. This allows the incorporation of physics-based residuals directly in the training loss, following the principles of physics-informed machine learning (PIML) \citep{raissiPhysicsinformedNeuralNetworks2019a}.




A key difficulty arises when enforcing geometry-dependent boundary conditions, such as the traction-free condition along the inner boundary representing the fracture. In the native DeepXDE implementation of DeepONet, boundary conditions are imposed regardless of the branch network inputs. To overcome this limitation, we extended the DeepONet framework to allow geometry-specific boundary conditions to be imposed in a weak, geometry-aware manner.

\begin{figure}[b]
        \centering
        \includegraphics[
    width=0.8\linewidth,
    trim = 0 0 0 14mm,
    clip
]{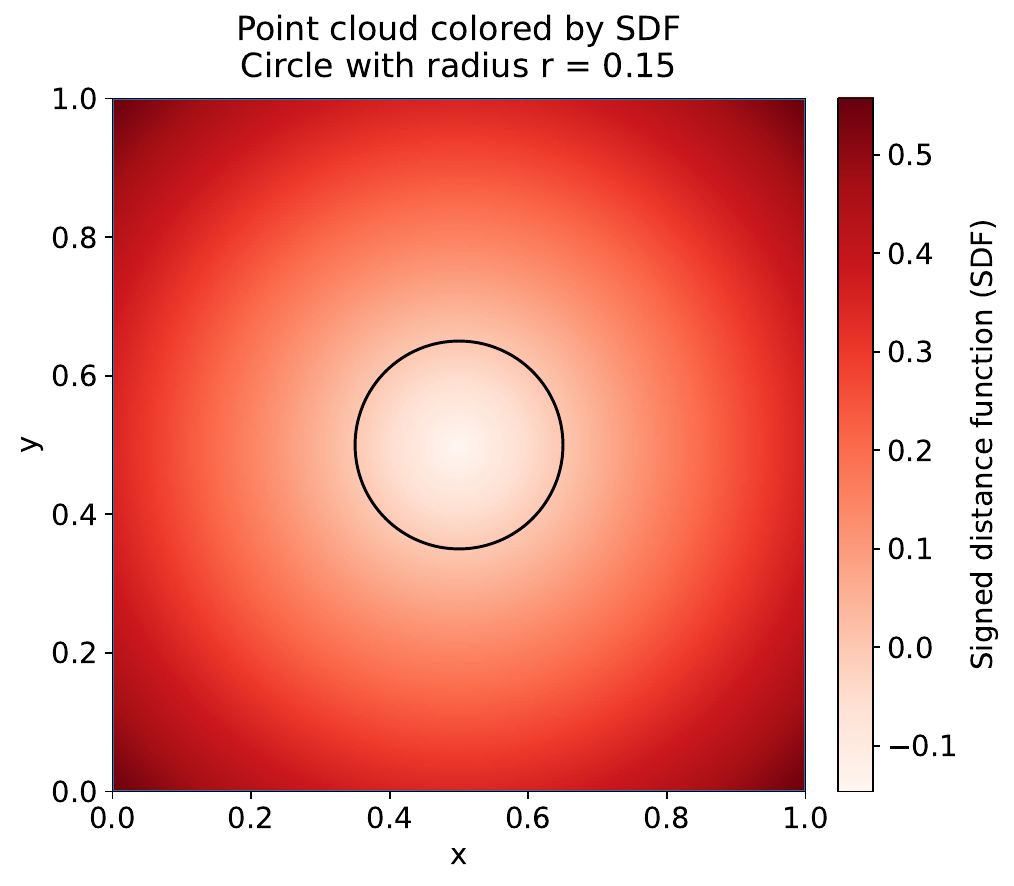}
        \caption{Signed Distance Function (SDF) to the circular hole boundary within the 2D domain.}
        \label{fig:sdf}
\end{figure}

To achieve this, we introduced a signed distance function, denoted by the symbol $\phi$, which measures the oriented distance from any evaluation point to the fracture boundary, as illustrated in Fig ~\ref{fig:sdf}. The value and gradient of $\phi$ provide local geometric information that enables the model to detect proximity to the fracture interface and enforce appropriate traction-free conditions.



In this work, two distinct strategies are proposed to leverage the signed distance function $\phi$ for imposing geometry-dependent conditions. Both approaches allow the operator to explicitly rely on the underlying geometry encoded by the branch network. This addresses the limitations of standard DeepONet formulations when dealing with problems involving internal boundaries.
\subsubsection{Option 1: Diffuse boundary imposition.}
In this first formulation, the signed distance function $\phi$ is used to enforce the traction-free condition along the internal boundary in a diffuse, geometry-aware manner. The learning process is guided by the following system of constraints:
\begin{equation}\label{eq:fracture-system}
\begin{aligned}
\begin{cases}
\nabla \cdot \boldsymbol{\sigma}(\mathbf{u}) = 0, & \text{in } \Omega, \\[8pt]
\delta(\phi)\,\boldsymbol{\sigma}(\mathbf{u})\,\mathbf{n} = 0, & \text{(diffuse free fracture)}, \\[8pt]
\mathbf{u} = 0, & \text{on fixed boundaries}, \\[8pt]
\boldsymbol{\sigma}(\mathbf{u})\,\mathbf{n}_b = \mathbf{t}_N, & \text{on traction boundaries.}
\end{cases}
\end{aligned}
\end{equation}

As reflected in the governing Eq. (\ref{eq:fracture-system}), a zero-divergence $\nabla \cdot \boldsymbol{\sigma}(\mathbf{u}) = 0$ condition is imposed on the stress field at interior points to enforce mechanical equilibrium. A clamped boundary condition is applied to the left edge $\mathbf{u} = 0$, prescribing zero displacement, while a normal traction load is imposed to the right edge $\boldsymbol{\sigma}(\mathbf{u})\,\mathbf{n}_b = \mathbf{t}_N$.


The orientation of the distance function gradient is provided by the direction of the shortest path to the fracture boundary from any point in the domain. Consequently, the gradient of the distance function, denoted by $ \nabla\phi$, corresponds to the normal vector to the fracture contour. A zero-traction condition, i.e. a zero normal stress condition, must be imposed along this boundary.



Using this gradient, the inner boundary normal can be retrieved and as a consequence, the traction-free condition can be enforced. This occurs in a diffuse manner through a weighting coefficient,  $\delta(\phi)$, which allows the network to utilize its interpolating capability up to the circular interface, where zero normal traction must strictly hold. When points are sampled near this interface, the penalty activates smoothly according to the value of $\delta(\phi)$.  To illustrate this concept, Eq. (\ref{eq:projection}) defines the projection of a point $p$ onto the fracture where ${\phi = 0}$. It moves $p$ along the direction of $\nabla \phi(p)$ by exactly the amount needed to reach the fracture contour, so that $p^{*}$ lies on the zero–level set where the traction-free boundary condition is imposed.
\begin{equation}
    p^{*} = p - \frac{\phi(p)}{\|\nabla \phi(p)\|^2} \, \nabla \phi(p)
    \label{eq:projection}
\end{equation}



\subsubsection{Option 2: Compliant material approach.}

The second approach for representing the internal boundary avoids enforcing the traction-free condition explicitly. Instead, the hole is modeled as a soft (or ``compliant'') material region whose mechanical properties differ significantly from those of the surrounding bulk. The signed distance function $\phi$ is again used to modulate these properties and to embed the geometric information directly into the governing equations.


\begin{equation}\label{eq:ersatz-system-H}
\left\{
\begin{aligned}
\nabla\cdot\boldsymbol{\sigma}_H(\mathbf{u}) &= 0,\\[4pt]
\boldsymbol{\sigma}_H(\mathbf{u})
&= H_\varepsilon(\phi)\,\boldsymbol{\sigma}_{\mathrm{b}}(\mathbf{u})
 + \bigl(1 - H_\varepsilon(\phi)\bigr)\,
   \boldsymbol{\sigma}_{\mathrm{s}}(\mathbf{u}),\\[4pt]
\mathbf{u} &= 0,\\[4pt]
\boldsymbol{\sigma}_H(\mathbf{u})\,\mathbf{n}_b &= \mathbf{t}_N.
\end{aligned}
\right.
\end{equation}

Here $\nabla\cdot\boldsymbol{\sigma}_H(\mathbf{u}) = 0$ and
$\boldsymbol{\sigma}_H(\mathbf{u})
= H_\varepsilon(\phi)\,\boldsymbol{\sigma}_{\mathrm{b}}(\mathbf{u})
 + (1 - H_\varepsilon(\phi))\,\boldsymbol{\sigma}_{\mathrm{s}}(\mathbf{u})$
hold in $\Omega$, the condition $\mathbf{u}=0$ is on fixed
boundaries, and
$\boldsymbol{\sigma}_H(\mathbf{u})\,\mathbf{n}_b = \mathbf{t}_N$
on traction boundaries.

In the implementation, we define a smooth indicator function of the intact material
\(
H_\varepsilon(\phi) = \tfrac{1}{2}\bigl(1 + \tanh(\phi / \varepsilon_{\mathrm{trans}})\bigr),
\)
so that \(H_\varepsilon(\phi) \approx 1\) in the bulk region (\(\phi > 0\)) and
\(H_\varepsilon(\phi) \approx 0\) inside the (diffuse) fracture region (\(\phi < 0\)).

To encode the fracture geometry as input for the branch network of the DeepONet, the distance function is evaluated at various points within the domain. Each geometry must be assigned a distinct encoding to ensure an injective mapping. Therefore, the number of evaluations and branch net entries must at least guarantee this. In the simple case of a circles with the same center, their radius distinguish them. Therefore only one additional point of evaluation besides the center is required to distinguish any number of different circles. Nevertheless, the branch network has the potential to convey richer geometric information than just differentiating between geometries; however, this topic lies outside the scope of the current study.




For the present work, we restrict attention to simple geometries and minimal encodings. The longer-term objective is to develop a more general formulation capable of handling arbitrary shapes and more complex geometric parameterizations, thus enabling the DeepONet to learn operator mappings over broader classes of domains.




\begin{figure}[ht]
        \centering
        \includegraphics[
    width=0.7\linewidth,
    trim = 0 0 0 12mm,
    clip
]{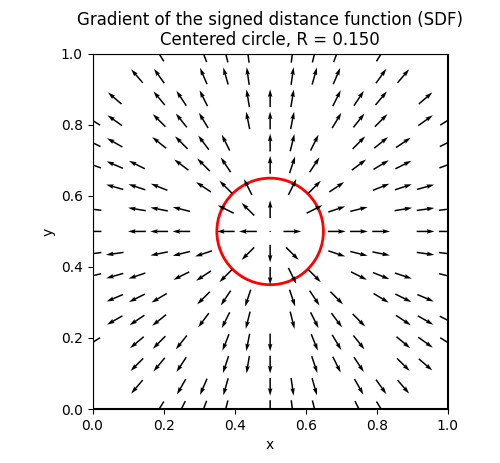}
        \caption{Example of SDF gradient field within 2D domain with a circle.}
        \label{fig:grad_sdf}
\end{figure}

\subsection{Finite Element Methods as a reference}

To validate the model output, it is compared with FEM simulation with identical geometry, loading, boundary conditions, and isotropy. The FEM simulation was performed with the FreeFem++ open source library \cite{MR3043640}.

The weak formulation of the linear elasticity problem solved by \textsc{FreeFem++} is given by:

\[
\int_{\Omega} \left( 
\lambda\, (\nabla\!\cdot u)(\nabla\!\cdot v)
\;+\;
2\mu\, \varepsilon(u) : \varepsilon(v)
\right)\, dx
=
\int_{\Gamma_N} T_0 \, v_x \, ds,
\]
where Lamé coefficients $\lambda$ and $\mu$ characterize an isotropic material: $\lambda$ controls the compressive response, whereas $\mu$ is the shear modulus and governs the deviatoric deformation. 
The tensor $\varepsilon(u)$ is the strain tensor. \(\Gamma_N\) is the right boundary.

\begin{figure}[ht]
    \centering
    \includegraphics[
    width=0.7\linewidth,
    trim = 0 0 0 8mm,
    clip
]{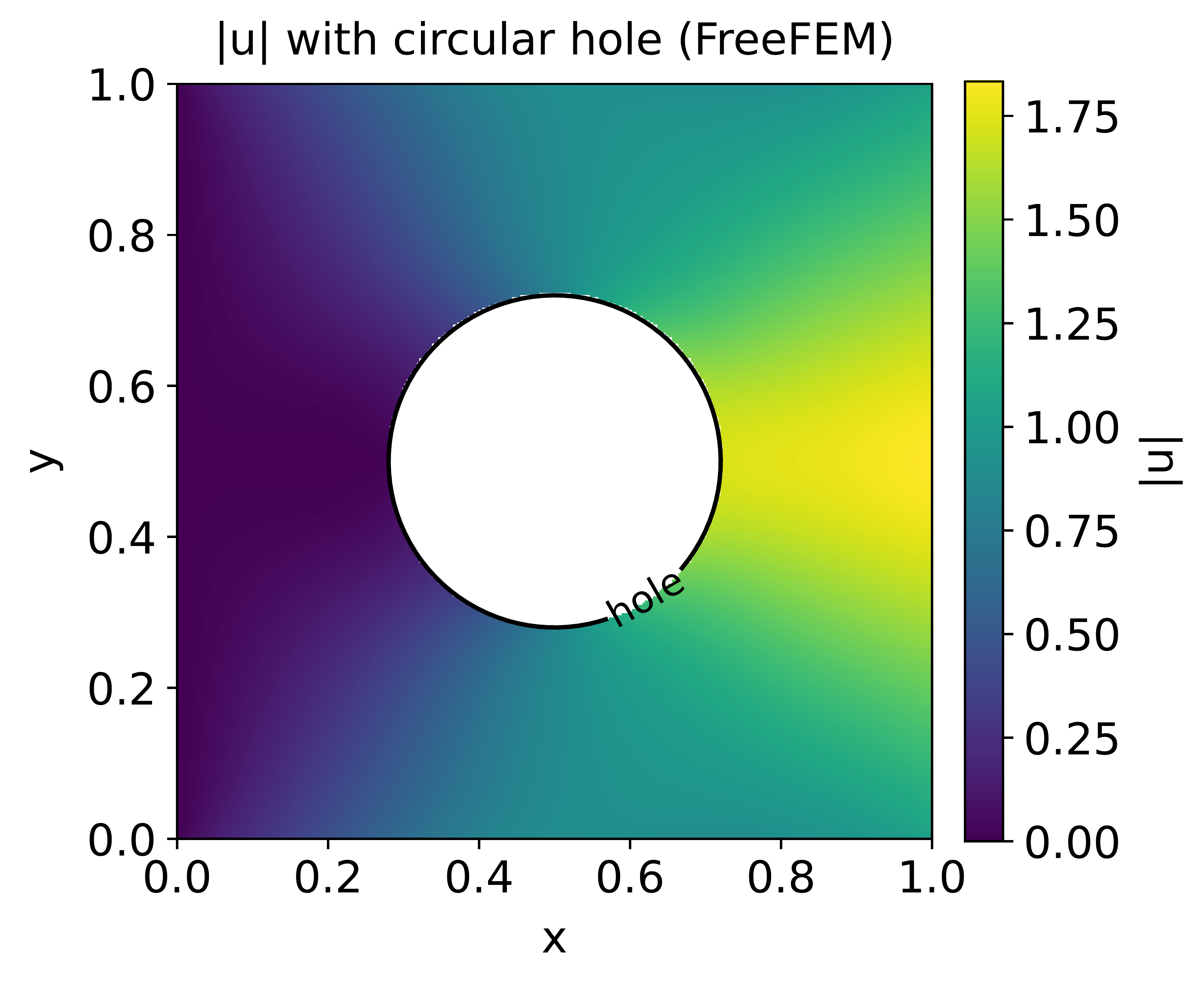}
    \caption{Norm of the displacement field over a domain with circular hole obtained by the FEM Simulation.}
    \label{fig:fem}
\end{figure}

The resulting displacement magnitude $\sqrt{u_x^2 + u_y^2}$ for 2D domain with a circular hole at its center is shown in Fig.~\ref{fig:fem}. For this simulation the meshing consists of $19{,}205$ nodes with a refined mesh along the inner circle boundary. This result is consistent with what can be found in the literature, for instance in \cite{jiangInvestigationEffectsVoids2014}.

\section{Results}


The learning process consists of evaluating each of the two options presented above over a grid of hyperparameters, including the learning rate, smoothing parameters, and sampling strategies. For the present study, the network depth was kept fixed, as was the number of collocation points: $20{,}000$ points were sampled in the domain and $5{,}000$ points on each boundary. The Lamé parameters were selected arbitrarily, $\lambda = 1$ and $\mu = 0.3$, without the intention of representing any specific physical material.


At this stage, only a single geometry with a fixed inner radius was considered. Consequently, no claim can be made regarding the generalization capability of the neural operator across multiple geometries.

The first formulation, which seeks to impose the traction-free condition explicitly through a diffuse penalty, did not converge to a physically meaningful solution. In the most favorable cases, the predicted displacement field corresponds essentially to that of a domain without any hole. The losses associated with the right-boundary traction and the weakly enforced traction-free constraint do not appear to balance during training, with one consistently dominating the other. Moreover, the sharp transition near the interface induces steep gradients. This target behavior makes it challenging for the network to accurately interpolate between the internal mechanical equilibrium and the traction-free boundary condition.

\begin{figure}[ht]
    \centering
    \includegraphics[
    width=0.735\linewidth,
    trim = 0 0 0 8mm,
    clip
]{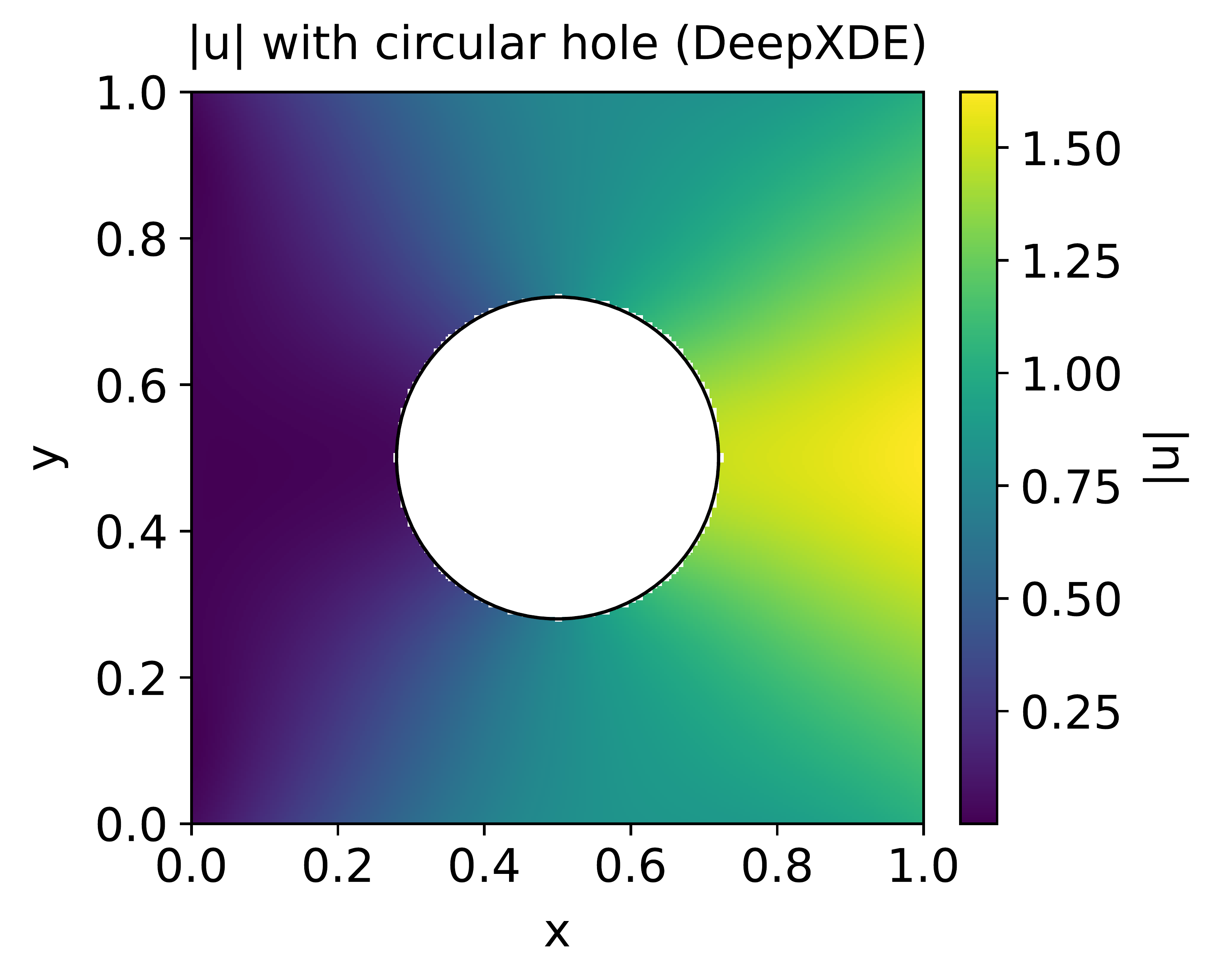}
    \caption{ Norm of the displacement field over a domain with circular hole obtained by the DeepONet surrogate model.}
    \label{fig:pinn}
\end{figure}
\begin{figure}[ht]
    \centering
    \includegraphics[
    width=0.8\linewidth,
    trim = 0 0 0 8mm,
    clip
]{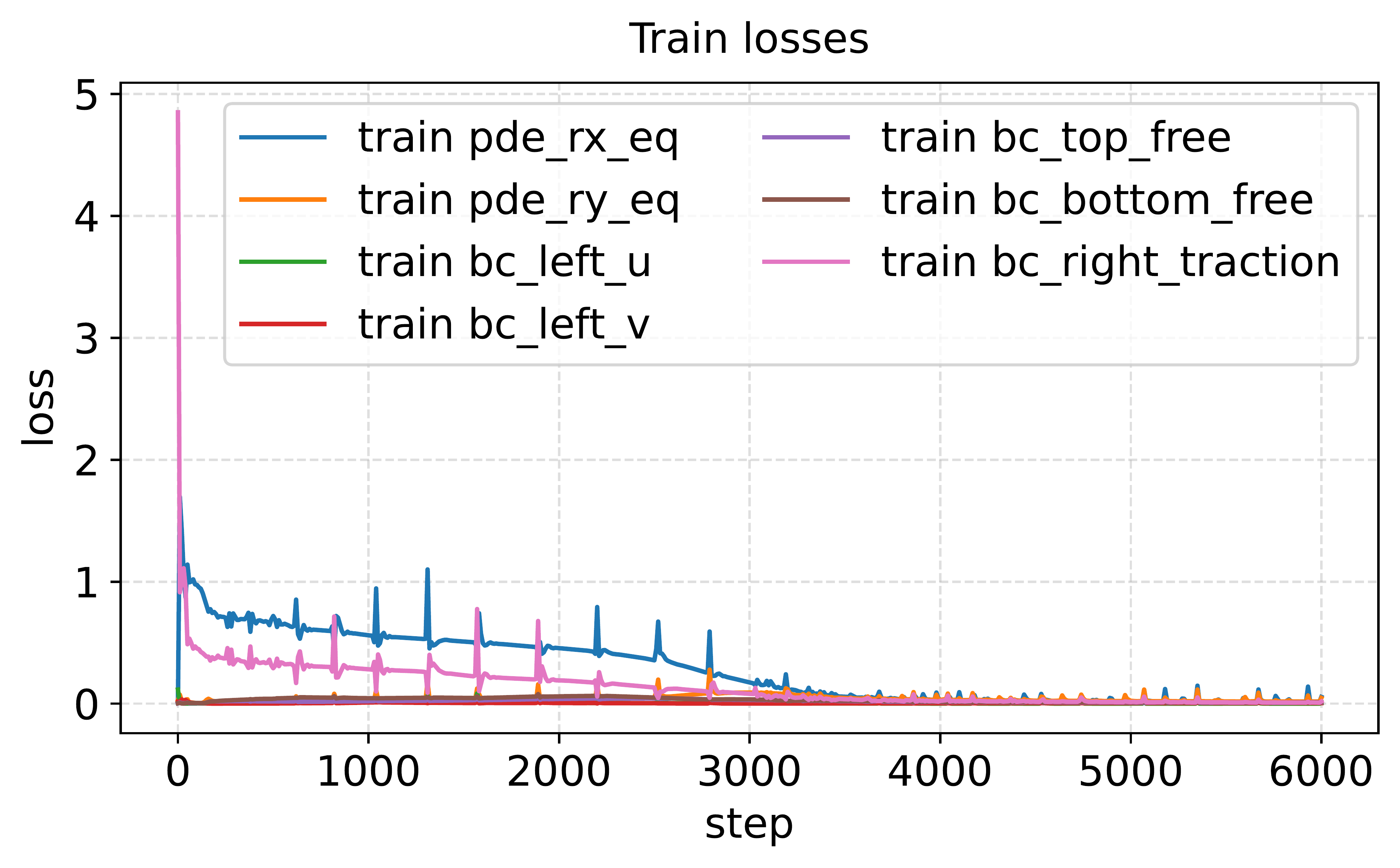}
    \caption{Evolution of the training losses. }
    \label{fig:losses}
\end{figure}

In contrast, the second option---in which the hole is modeled as a soft material---is the one that delivered coherent results. The training converged, as shown in Fig.~\ref{fig:losses}, and it is physically coherent. This is the option for which the results are illustrated hereafter. The corresponding displacement magnitude, illustrated in Fig.~\ref{fig:pinn}, exhibits a pattern qualitatively consistent with the FEM reference solution, with discrepancies on the order of a few decimal digits. For this visualization, the hyperparameter configuration producing the lowest training loss was selected empirically, although several other tested configurations yielded similarly satisfactory convergence.


\begin{figure}[ht]
    \centering
    \includegraphics[
    width=0.8\linewidth,
    trim = 0 0 0 6.6mm,
    clip
]{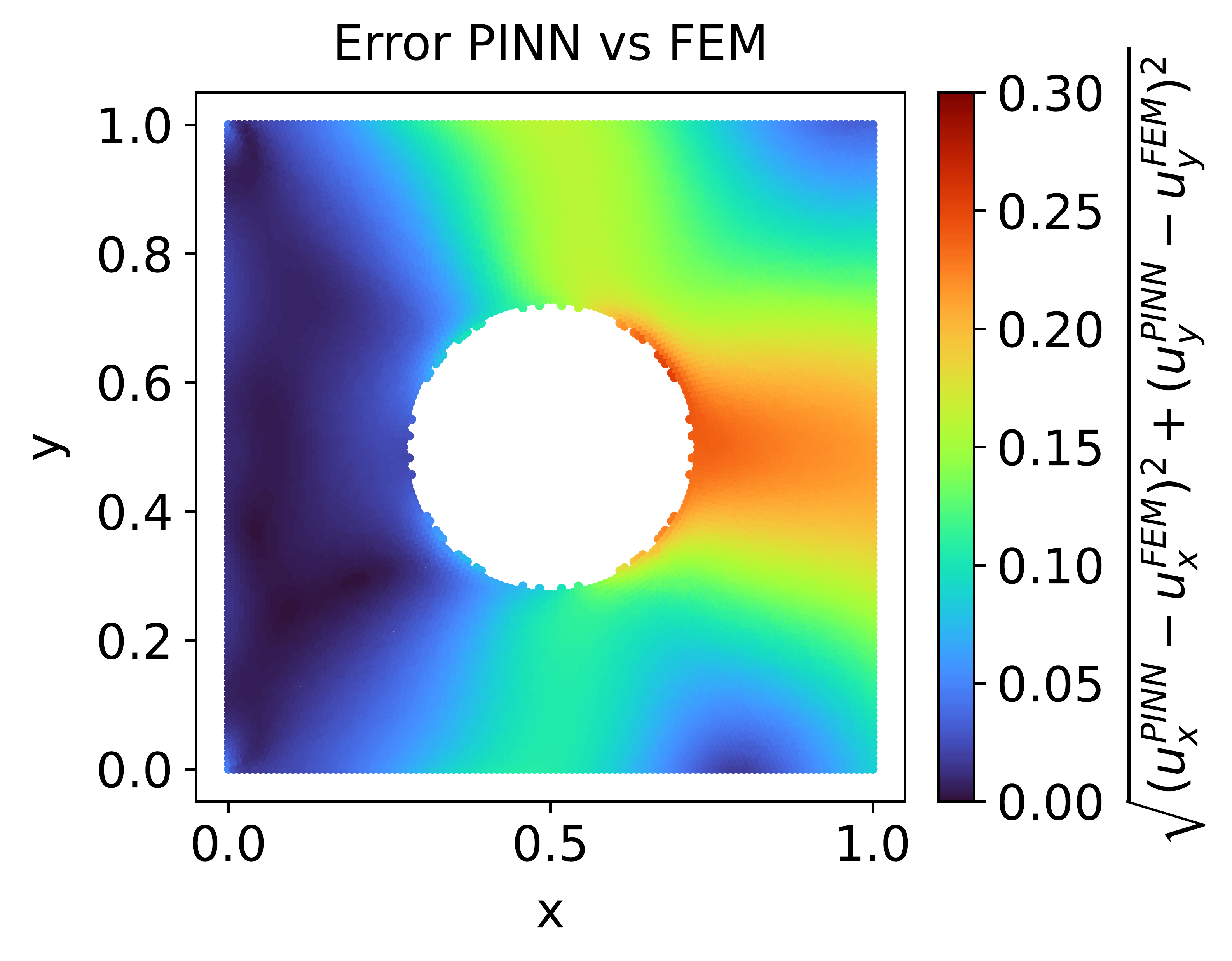}
    \caption{Error between FEM and the physics-based learning surrogate model.}
    \label{fig:error}
\end{figure}

Fig.~\ref{fig:error} shows the pointwise error, evaluated on the same set of sample points for both the finite element solution and the DeepONet model.

The error is concentrated in regions where large displacements are expected, where the traction imposed on the right boundary competes with the (nearly) zero traction implicitly enforced on the inner circular boundary. This is most likely due to the function $H_{\varepsilon}(\phi)$ that smooths the transition between the approximated hole and the rest of the domain. It is precisely at this point that the approximation is introduced, and it is also where geometric singularities, and therefore stress concentrations, arise.

\section{Conclusion and perspectives}





The present work demonstrates the feasibility of using a physics-informed DeepONet framework to predict displacement fields in perforated elastic domains without relying on FEM-generated training data. We investigated two variants to train the network to reconstruct the field of displacement depending on a domain geometry under a tensile load. Several directions for improvement and extension emerge from these results.

A first avenue concerns the architecture of the trunk network, which governs how spatial coordinates contribute to the operator output. More expressive architectures may allow the model to better capture the geometric specificities of the domain. For example, Kolmogorov--Arnold Networks (KANs)~\citep{liuKANKolmogorovArnoldNetworks2025}, already used effectively for fracture prediction in~\cite{kiyaniPredictingCrackNucleation2025}, could enhance the representation of spatially complex solution fields. Similarly, improvements can be made in the sampling procedure: oversampling the region around the fracture, in the spirit of local mesh refinement in FEM, may reduce errors in regions dominated by steep gradients or stress concentrations.


The present study only considers a simple two-dimensional configuration that does not fully reflect realistic fracture scenarios. It remains to investigate whether the proposed operator-learning approach can generalize to more complex geometries—such as non-circular defects, arbitrary shapes, or three-dimensional domains—and to assess the associated computational cost. Moreover, because the network enforces continuity to facilitate convergence, it may struggle to reproduce features associated with geometric singularities. 
For instance, while the analytical solution exhibits stress concentrations at the corners of a rectangular domain, the localized smoothing introduced by our formulation can prevent the model from capturing these sharp variations.
A more systematic analysis of the smoothing procedure and its influence on accuracy in critical regions is needed.



\begin{figure}[t]
    \centering
    \includegraphics[width=0.7\linewidth]{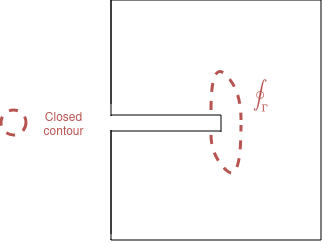}
    \caption{$\Gamma$ is any curve surrounding the notch tips as described in (Rice, 1968).}
    \label{fig:j-integral}
\end{figure}

A longer-term objective of this line of work is to infer fracture propagation directly from the displacement fields predicted by the neural operator. In this context, the Paris--Erdogan law provides an empirical relationship between the fracture growth rate and the range of the stress intensity factor,
\[
\Delta K = K_{\max} - K_{\min}, \qquad 
\frac{da}{dN} = C (\Delta K)^m,
\]
where $C$ and $m$ are material-dependent constants. To evaluate $\Delta K$ without relying on FEM, one may compute the $J$-integral~\citep{ricePathIndependentIntegral1968} from the predicted displacement field,
\[
J = \int_{\Gamma} \left( 
W\, \delta_{1j} 
- 
\sigma_{ij} \frac{\partial u_i}{\partial x_1}
\right) n_j \, ds ,
\]
with each term expressed in terms of displacement gradients and Hooke's law. Fig.~\ref{fig:j-integral} illustrates a representative contour $\Gamma$ surrounding the defect (see \cite{ricePathIndependentIntegral1968}). Since this method is designed for sharp cracks, the neural operator must ultimately be able to represent displacement fields in geometries with discontinuous boundaries and strong stress singularities—an important challenge for future developments.


\begin{ack}

This work benefited from the GPU computing resources provided by the LAAS laboratory in Toulouse (France).
\end{ack}

\section*{DECLARATION OF GENERATIVE AI AND AI-ASSISTED TECHNOLOGIES IN THE WRITING PROCESS}
During the preparation of this work the authors used ChatGPT in order to translate some parts in English for fluency. After using this tool/service, the author reviewed and edited the content as needed and take full responsibility for the content of the publication.

\bibliography{ifacconf}             

@article{raissiPhysicsinformedNeuralNetworks2019a,
	title = {Physics-informed neural networks: {A} deep learning framework for solving forward and inverse problems involving nonlinear partial differential equations},
	volume = {378},
	issn = {0021-9991},
	shorttitle = {Physics-informed neural networks},

	journal = {Journal of Computational Physics},
	author = {Raissi, M. and Perdikaris, P. and Karniadakis, G. E.},
	month = feb,
	year = {2019},
	pages = {686--707},
}

@article{kiyaniPredictingCrackNucleation2025,
	title = {Predicting crack nucleation and propagation in brittle materials using {Deep} {Operator} {Networks} with diverse trunk architectures},
	volume = {441},
	issn = {0045-7825},

	journal = {Computer Methods in Applied Mechanics and Engineering},
	author = {Kiyani, Elham and Manav, Manav and Kadivar, Nikhil and De Lorenzis, Laura and Karniadakis, George Em},
	month = jun,
	year = {2025},
	pages = {117984},
}

@article{jiangInvestigationEffectsVoids2014,
	title = {An investigation into the effects of voids, inclusions and minor cracks on major crack propagation by using {XFEM}},
	volume = {49},
	journal = {Structural Engineering and Mechanics},
	author = {Jiang, Shouyan and Du, Chengbin and Gu, Chongshi},
	month = mar,
	year = {2014},
}

@article{liuKANKolmogorovArnoldNetworks2025,
	title = {{KAN}: {Kolmogorov}-{Arnold} {Networks}},
	shorttitle = {{KAN}},
	journal = {arXiv},
	author = {Liu, Ziming and Wang, Yixuan and Vaidya, Sachin and Ruehle, Fabian and Halverson, James and Soljačić, Marin and Hou, Thomas Y. and Tegmark, Max},
	month = feb,
	year = {2025},
}

@article{drosopoulosDeepONetPredictionFailure2025,
	title = {{DeepONet} for the {Prediction} of {Failure} {Response} of a {Two}-{Dimensional} {Fibre}-{Reinforced} {Composite} {Plate}},
	volume = {1},
	issn = {3083-4643, 3083-4643},
	language = {en},
	number = {1},

	journal = {Bulletin of Computational Intelligence},
	author = {Drosopoulos, Georgios A. and Stavroulakis, Georgios E. and Drosopoulos, Georgios A. and Stavroulakis, Georgios E.},
	month = sep,
	year = {2025},

	pages = {76--88},
}

@book{brennerMathematicalTheoryFinite2008,
	address = {New York, NY},
	series = {Texts in {Applied} {Mathematics}},
	title = {The {Mathematical} {Theory} of {Finite} {Element} {Methods}},
	volume = {15},
	copyright = {http://www.springer.com/tdm},
	isbn = {978-0-387-75933-3 978-0-387-75934-0},
	publisher = {Springer},
	author = {Brenner, Susanne C. and Scott, L. Ridgway},
	year = {2008},
}

@article{luDeepONetLearningNonlinear2021,
	title = {{DeepONet}: {Learning} nonlinear operators for identifying differential equations based on the universal approximation theorem of operators},
	volume = {3},
	issn = {2522-5839},
	shorttitle = {{DeepONet}},
	language = {en},
	number = {3},

	journal = {Nature Machine Intelligence},
	author = {Lu, Lu and Jin, Pengzhan and Karniadakis, George Em},
	month = mar,
	year = {2021},
	pages = {218--229},
}

@article{ricePathIndependentIntegral1968,
	title = {A {Path} {Independent} {Integral} and the {Approximate} {Analysis} of {Strain} {Concentration} by {Notches} and {Cracks}},
	volume = {35},
	issn = {0021-8936},
	number = {2},

	journal = {Journal of Applied Mechanics},
	author = {Rice, J. R.},
	month = jun,
	year = {1968},
	pages = {379--386},
}

@article{baydinAutomaticDifferentiationMachine2018a,
  title={Automatic differentiation in machine learning: a survey},
  author={Baydin, Atilim Gunes and Pearlmutter, Barak A and Radul, Alexey Andreyevich and Siskind, Jeffrey Mark},
  journal={Journal of machine learning research},
  volume={18},
  number={153},
  pages={1--43},
  year={2018}
}

@article{MR3043640,

  AUTHOR = {Hecht, F.},
  TITLE = {New development in FreeFem++},
  JOURNAL = {J. Numer. Math.},
  FJOURNAL = {Journal of Numerical Mathematics},
  VOLUME = {20}, YEAR = {2012},
  NUMBER = {3-4}, PAGES = {251--265},
  ISSN = {1570-2820},
  MRCLASS = {65Y15},
  MRNUMBER = {3043640},
  URL = {https://freefem.org/}

}

@article{ciklamini2025enhancing,
  title={Enhancing digital twin performance through optimizing graph reduction of finite element models},
  author={Ciklamini, Marek and Cejnek, Matous},
  journal={Scientific Reports},
  volume={15},
  number={1},
  pages={37777},
  year={2025},
  publisher={Nature Publishing Group UK London}
}
                                                   







\appendix

\end{document}